%% file: mtsummit25.tex
\pdfoutput=1

\documentclass[11pt]{article}

\usepackage[final]{mtsummit25}
\usepackage{copyright}

\usepackage{times}
\usepackage{latexsym}

\usepackage[T1]{fontenc}

\usepackage[utf8]{inputenc}

\usepackage{microtype}

\usepackage{inconsolata}

\usepackage{graphicx}

\usepackage[british]{babel}
\usepackage[babel]{csquotes}
\usepackage{booktabs}
\usepackage{tabularx}
\PassOptionsToPackage{hyphens}{url}\usepackage{hyperref}

\newcommand{\Section}[1]{Section~\ref{sec:#1}}
\newcommand{\Table}[1]{Table~\ref{tab:#1}}

\title{A comparison of translation performance between DeepL and Supertext}

\author{
 \parbox{0.8\linewidth}{\centering
     \textbf{Alex Flückiger},
     \textbf{Chantal Amrhein},
     \textbf{Tim Graf},
     \textbf{Frédéric Odermatt},
     \textbf{Martin Pömsl},
     \textbf{Philippe Schläpfer},
     \textbf{Florian Schottmann},
     \textbf{Samuel Läubli}}
     \vspace{0.6em}
\\
 Supertext
 \vspace{0.6em}
\\
 \texttt{\{firstname.lastname\}@supertext.com}
}

\begin{document}
\maketitle 
\begin{abstract}
As strong machine translation (MT) systems are increasingly based on large language models (LLMs), reliable quality benchmarking requires methods that capture their ability to leverage extended context. This study compares two commercial MT systems -- DeepL and Supertext -- by assessing their performance on unsegmented texts. We evaluate translation quality across four language directions with professional translators assessing segments with full document-level context. While segment-level assessments indicate no strong preference between the systems in most cases, document-level analysis reveals a preference for Supertext in three out of four language directions, suggesting superior consistency across longer texts. We advocate for more context-sensitive evaluation methodologies to ensure that MT quality assessments reflect real-world usability.\footnote{Paper accepted at MT Summit 2025.}\footnote{We release all evaluation data and scripts for further analysis and reproduction at \url{https://github.com/supertext/evaluation_deepl_supertext}}
\end{abstract}

\section{Introduction}
\label{sec:Introduction}
After the transition from statistical to neural modelling roughly a decade ago 
\citep{Kalchbrenner2013,Sutskever2014,Bahdanau2015}, the field of MT is undergoing another paradigm shift towards leveraging LLMs \citep{xu2024paradigmshiftmachinetranslation,wu2024adaptinglargelanguagemodels,kocmi-etal-2024-findings}. LLM-based translation offers the potential for significantly improved translation quality, especially with respect to consistent translation of documents. Unlike neural machine translation (NMT) systems, which typically process documents as isolated sentences or paragraphs \citep{Post2023EscapingTS}, many LLMs operate with context windows that can span thousands of words, allowing them to maintain consistency throughout a document -- for instance, by ensuring that a word’s translation in the final sentence matches its previous forms \citep{wu2024adaptinglargelanguagemodels}.

In the most recent shared task at the Conference on Machine Translation (WMT24) that focuses on evaluating the state of the art in general-domain translation quality, the majority of the 28 system submissions were already based on LLMs \citep{kocmi-etal-2024-findings}. 
Although without statistical significance and for the language direction English to German only, one system even outranked the human reference translations as evaluated by professional human annotators.

Despite this impressive achievement, findings of human-machine parity should be approached with caution. Similar claims already emerged with pre-LLM technology \citep{hassan2018achieving}, yet have subsequently been refuted due to limitations in the evaluation design focusing on single segments in isolation \citep{laubli-etal-2018-machine,toral-etal-2018-attaining, freitag-etal-2021-experts}. The WMT24 shared task also highlights that evaluations based on automatic metrics (rather than human evaluation) can lead to wrong conclusions when comparing strong MT systems \citep{kocmi-etal-2024-findings}.

However, these insights are often overlooked in evaluations of commercial MT systems. For example, Intento's The State of Machine Translation 2024 report,\footnote{\label{Intento}\url{https://inten.to/machine-translation-report-2024}} which assesses 52 providers across 11 language pairs, serves as a valuable resource for potential users in real-world settings, but its benchmarking methodology relies on automatic scoring of sentence-level data, and the authors acknowledge that \enquote{you may need a human linguist} to ensure greater reliability.

In this paper, we evaluate two commercial translation systems (\Section{Systems}) under conditions that allow for leveraging the full-text capabilities of LLMs. The segmentation of the source text is handled by the translation systems alone without any prior splitting (\Section{Data}), and the resulting translations are rated by professional translators considering the full document context (\Section{EvaluationSetup}). We find that while both systems translate a similar number of segments better than the other, the difference is more pronounced on the document level (\Section{Results}), which we attribute to differences in how much context the systems consider during translation (\Section{Discussion}). Our findings suggest that the adoption of LLMs creates opportunities for smaller players to challenge dominant industry leaders (\Section{Conclusion}).

\section{Systems}
\label{sec:Systems}

We compare the free online offering of two commercial MT providers:

\paragraph{DeepL}
DeepL\footnote{\url{https://www.deepl.com}} is a widely used MT provider boasting \enquote{unrivalled translations that set the standard}.\footnote{\label{DeepLQuality}\url{https://www.deepl.com/en/quality}, see also Appendix~\ref{sec:Appendix}.} In the latest Intento report, it scores best among nine \enquote{real-time engines} and, together with GPT-4, is found to \enquote{consistently outperform other models}.\footref{Intento} Due to its closed source, the technology behind DeepL's translation system is not publicly known.

\paragraph{Supertext} 
Supertext\footnote{\url{https://www.supertext.com}} builds on an open, general-purpose LLM that has been specialised for the task of translation with proprietary methods and data. While the system can be adapted to specific domains, we use the freely available generic version.


\noindent For the purpose of the evaluation described in this paper, we use both systems with default settings. For example, we do not specify politeness (formal/informal) although supported by both systems in some language combinations.

While both DeepL and Supertext provide target language variants for English (British and American), Supertext also provides target language variants for German (Austria, Germany, Switzerland), French (France, Switzerland), and Italian (Italy, Switzerland). As our use case is machine translation for people in Switzerland, we use the Swiss target language variant whenever available (\Section{TargetTexts}).\footnote{Compared to English variants, the Swiss variants of other languages differ minimally.}

\section{Data}
\label{sec:Data}

\subsection{Source Texts}
\label{sec:SourceTexts}

We collect 20 texts each in two source languages: English (en) and German (de). All texts stem from news websites: New York Times\footnote{\url{https://www.nytimes.com}} for English and Neue Zürcher Zeitung\footnote{\url{https://www.nzz.ch}} for German, respectively. We select 10 FAQ pages and 10 recent news articles in the economy section from each website. Notably, these texts are only available in a single language; they are unlikely to be contained in the training data of either system we evaluate. To balance the distribution of text lengths, we trim the end of some texts by omitting their final paragraphs. 

\subsection{Target Texts}
\label{sec:TargetTexts}

We create translations in four language directions (\Table{data_stats}) directly in the respective online translation interface of each system as a regular user would.\footnote{All translations were produced on 27 January 2025.} We do not modify the texts before translation and paste them in their original formatting, including newlines. The translation systems may segment the text into smaller chunks internally.

After translation, we manually split and align the source texts and translations into sentences. If one of the systems merges two or more sentences into a single sentence, we ensure that the same content is merged for the other system, such that the raters are presented with parallel segments. Table \ref{tab:data_stats} shows the resulting number of segments per language pair. The texts per source language are identical, differing only in how they were manually segmented for the A/B test after translation. Across the language pairs, the median number of segments per document is 13.

\begin{table}[t]
    \centering
    \begin{tabular}{lrrr}
        \toprule
        Language Direction & Texts & Segments & Words\\
        \midrule
        de $\rightarrow$ en-GB & 20 & 281 & 3336\\
        de $\rightarrow$ fr-CH & 20 & 276 & 3336\\
        de $\rightarrow$ it-CH & 20 & 265 & 3336\\
        en $\rightarrow$ de-CH & 20 & 211 & 3483\\
        \midrule
        Total & 80 & 1033 & 13491 \\
        \bottomrule
    \end{tabular}
    \caption{Evaluation data by language direction.}
    \label{tab:data_stats}
\end{table}

\section{Evaluation Setup}
\label{sec:EvaluationSetup}

We conduct a blind A/B test in which professional translators rate DeepL and Supertext outputs with full document-level context.

\subsection{Raters}
\label{sec:Raters}

We enrol 8 professional translators with experience in evaluating machine translation output, 1 to 3 per language direction. All raters have between 2 and 19 years of professional experience (average=8.6 years) in the language combination they are assigned to and are native in the respective target language.

\subsection{Materials}
\label{sec:Materials}

We arrange all segments of a source document with their corresponding translations by both systems in a spreadsheet. The segments are presented in original document order, including formatting such as newlines, such that raters see the full source text and both translations side-by-side. We randomly assign the system outputs to columns labelled Translation A and Translation B for each text such that raters do not have any information about which translation stems from which system (a blind A/B test setting). System assignments are kept consistent within a text such that the document context remains natural.

\subsection{Procedure}
\label{sec:Procedure}

Documents are assigned to single raters. For each segment in each document, the assigned rater is asked to choose whether Translation A is better, Translation B is better, or whether both translations are of equal quality.

Our instructions explicitly state that \enquote{equal} can mean that two translations are equally good or equally bad. Moreover, the raters were asked to focus on the content rather than punctuation to avoid that the results get biased because of specifics of a language variant.

\section{Evaluation Results}
\label{sec:Results}

\begin{figure}[t]
    \centering
    \includegraphics[height=4.48cm]{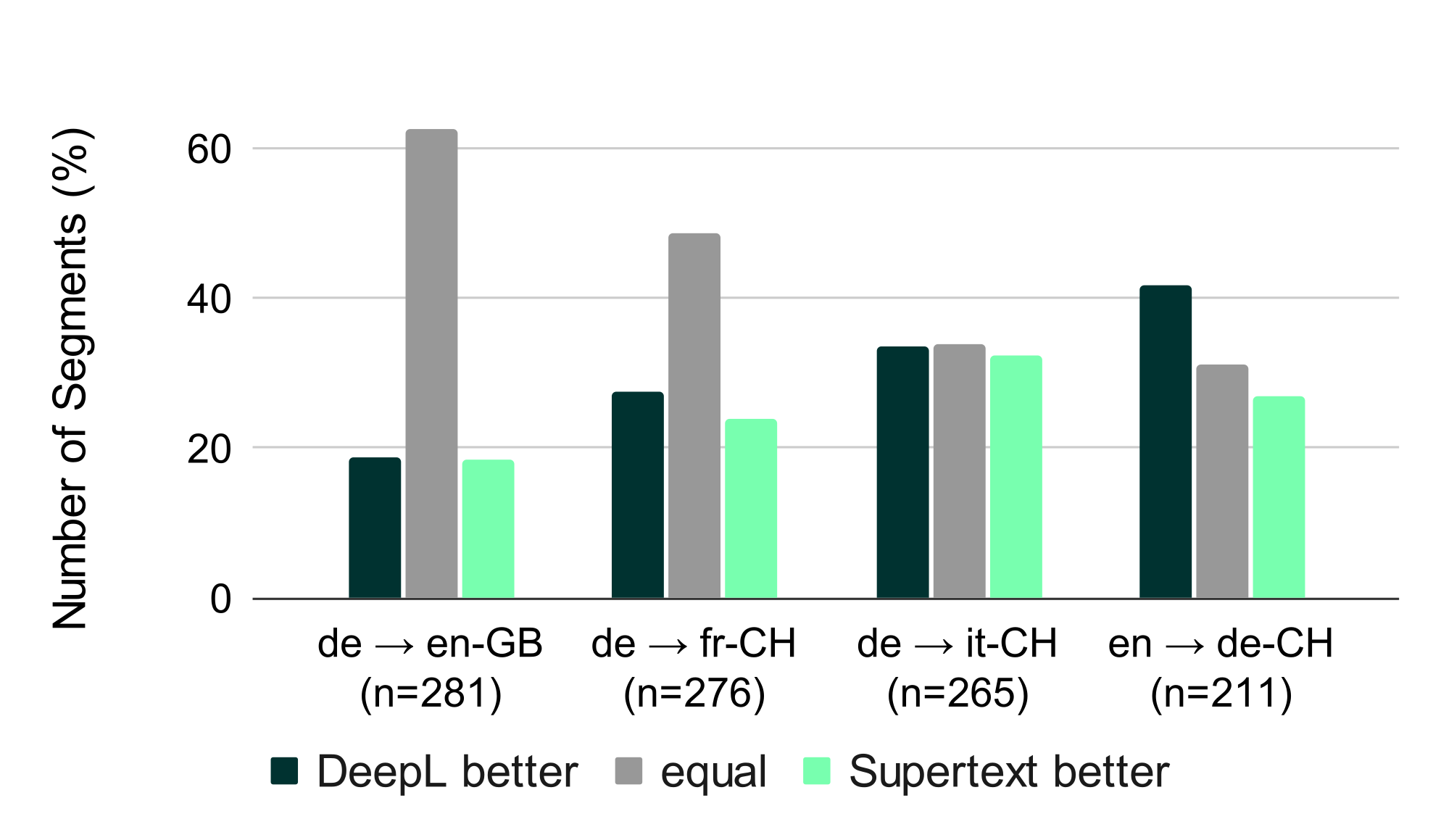}
    \caption{Segment-level ratings.}
    \vspace{2.3em}
    \label{fig:by-segment}
\end{figure}

\begin{figure}[t]
    \centering
    \includegraphics[height=4.48cm]{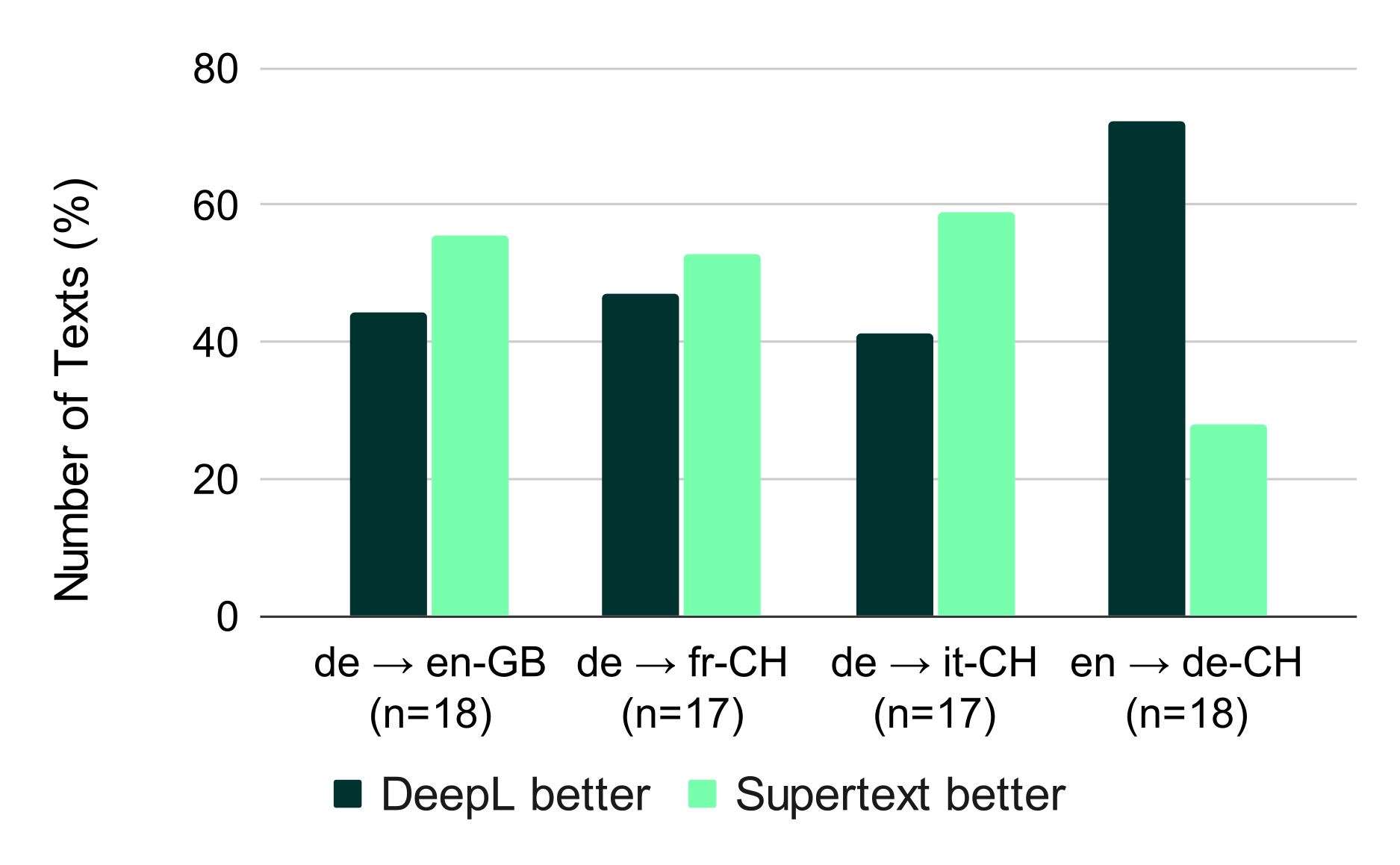}
    \caption{Aggregated segment-level ratings per text. Texts with the same number of preferred segments for both systems are excluded.}
    \label{fig:by-text}
\end{figure}

Segment-level and text-level preference ratings are shown in Figures~\ref{fig:by-segment} and~\ref{fig:by-text}, respectively.

\subsection{Segment-level}
\label{sec:ResultsSegmentLevel}
Across all language pairs, 9.5\% of the segments generated by DeepL and Supertext are identical. The overlap is highest in de $\rightarrow$ en-GB, particularly in the FAQ texts where 26.1\% of the segments were translated identically.

Participants rate most segments as equal in terms of translation quality in three out of four language directions. While the number of segments where one system is preferred over the other is similar for DeepL and Supertext in these language directions, raters show a preference for DeepL in en $\rightarrow$ de-CH (88~DeepL, 66~equal, 57~Supertext).

\subsection{Document-level}
\label{sec:ResultsDocumentLevel}
We derive document-level preferences by aggregating the segment-level ratings of each evaluated document. For example, a text is counted as \enquote{DeepL better} if the rater preferred DeepL's translations for more segments than Supertext's translations in that document.

In contrast to the pooled segment-level ratings (\Section{ResultsSegmentLevel}), raters show a preference for documents translated by Supertext in three out of four language directions, most notably in de $\rightarrow$ it-CH (7~DeepL, 3~equal, 10~Supertext). In en $\rightarrow$ de-CH, however, raters show a clear preference for documents translated by DeepL (13~DeepL, 2~equal, 5~Supertext).

\begin{table*}[t]
    \tiny
    \input{example}
    \caption{Example of a rated de $\rightarrow$ en-GB document. Better-rated translations are highlighted in bold; segments without any bold translation were rated as equal. System names are not shown during evaluation (\Section{EvaluationSetup}).}
    \label{tab:example}
\end{table*}

\section{Discussion}
\label{sec:Discussion}
Our evaluation highlights that conclusions drawn from MT quality assessments may vary significantly depending on the unit of measurement. While raters in our study preferred a similar number of translated segments by DeepL and Supertext overall, the difference becomes more pronounced at the document level. This discrepancy suggests that segment-level assessments alone may not fully capture translation quality as perceived in real-world usage, where coherence and consistency across entire documents play a critical role.

Notably, while segment-level ratings indicate no strong preference between the two systems in most language directions, document-level aggregation reveals a more distinct pattern. Raters favour Supertext's translations at the document level in three out of four language directions, with the most pronounced difference observed in de $\rightarrow$ it-CH. This suggests that Supertext may provide better consistency or fluency across longer texts in these language directions. While we have yet to conduct a systematic qualitative comparison of system outputs, we find texts where the same word is translated differently by DeepL and consistently by Supertext across paragraphs. An example is shown in \Table{example}, where DeepL translates the German word \textit{Startseite} as either \textit{start page}, \textit{home page}, or \textit{Home page}.

In contrast, for en $\rightarrow$ de-CH, raters show a clear preference for DeepL at both the segment and document levels, indicating a potential strength of DeepL in handling this specific language combination. Our preliminary analysis is inconclusive at this point, but the Supertext outputs seem to contain a higher number of within-sentence errors such as wrong choices for individual words or omissions. Another hypothesis is that Supertext, which supports three different German target language variants, may introduce inconsistencies by mixing region-specific elements in translation outputs.

\section{Conclusion}
\label{sec:Conclusion}

Our study highlights the growing significance of document-level evaluations in MT quality benchmarking, especially as LLM-based systems leverage broader context windows to enhance translation consistency. While segment-level assessments suggest no clear preference between DeepL and Supertext in most of the language directions we examined, document-level aggregation reveals notable differences. Supertext is preferred in three out of four language pairs, where its translations exhibit greater consistency. In contrast, en $\rightarrow$ de-CH shows a clear preference for DeepL, possibly due to fewer within-sentence errors or differences in regional language handling.

As LLM-based MT systems continue to evolve, future studies should further investigate the impact of context length on commercial MT benchmarking campaigns. Insights into how different systems leverage context and resolve ambiguities will be essential for advancing evaluation methodologies and ensuring that translation systems meet real-world user expectations.

\section*{Limitations}
While A/B tests are commonly used for comparing two systems and a reliable basis for incrementally improving MT systems \citep{tang_2010, wu-etal-2024-evaluating-automatic}, 
they provide no insight into the severity of errors within a translation or across different systems compared to MQM ratings \citep{freitag-etal-2021-experts}. Absent the use of more time-intensive evaluation frameworks, such limitations persist irrespective of whether preferences are aggregated at the system level or pooled by document.

During real-world usage, some mistakes may be harder to spot than others when not being shown contrastively against an alternative translation. Similarly, the preference in an A/B test may not correlate with the effort needed for post-editing the translation. To address these questions, we plan to extend our evaluation efforts.

The evaluation was carried out by professional translators working for Supertext. Since the A/B assignments were randomized and anonymized, we do not assume any bias. Additionally, in the interest of transparency, we publicly share the complete dataset, including the source text, translations from each system, and the corresponding ratings.

Finally, the scope of this study is not exhaustive but is limited to a subset of language pairs, two domains, and a limited number of documents. Yet, we are providing details that go beyond what DeepL is sharing publicly on their website.\footref{DeepLQuality}

\section*{Acknowledgments}
We thank all the professional translators involved for their support with this evaluation.

\bibliography{anthology,mtsummit25}

\appendix

\section{Appendix}
\label{sec:Appendix}

For the sake of persistency, we share the archived link as well:
\url{https://web.archive.org/web/20250215011944/https://www.deepl.com/en/quality}

\end{document}

%% file: example.tex
{\renewcommand{\arraystretch}{1.3}
\begin{tabularx}{\textwidth}{l>{\raggedright\arraybackslash}X>{\raggedright\arraybackslash}X>{\raggedright\arraybackslash}X}
\toprule
SID   & Source text (de)                                                                                                                               & DeepL (en-GB)                                                                                                                              & Supertext (en-GB)                                                                                                                              \\ \midrule
1  & Wie kann ich die NZZ als Startseite festlegen?                                                                                                    & How can I set the NZZ as my \underline{start page}?                                                                                                    & \textbf{How can I set NZZ as my \underline{homepage}?}                                                                                                     \\
   &                                                                                                                                                   &                                                                                                                                            &                                                                                                                                                \\
2  & Öffnen Sie Ihren Browser:                                                                                                                         & Open your browser:                                                                                                                         & Open your browser:                                                                                                                             \\
3  & - Stellen Sie sicher, dass der Browser geöffnet ist, den Sie verwenden möchten (z.B. Google Chrome, Mozilla Firefox, Microsoft Edge, Safari).     & - Make sure the browser you want to use is open (e.g. Google Chrome, Mozilla Firefox, Microsoft Edge, Safari).                             & - Make sure the browser you want to use is open (e.g., Google Chrome, Mozilla Firefox, Microsoft Edge, Safari).                                \\
   &                                                                                                                                                   &                                                                                                                                            &                                                                                                                                                \\
4  & Gehen Sie zu den Einstellungen:                                                                                                                   & Go to the settings:                                                                                                                        & Go to settings:                                                                                                                                \\
5  & - In den meisten Browsern finden Sie die Einstellungen oder Optionen im Menü oben rechts, oft dargestellt durch drei Punkte (?) oder Striche ($\equiv$). & - In most browsers, you will find the settings or options in the menu at the top right, often represented by three dots (?) or dashes ($\equiv$). & - In most browsers, you'll find the settings or options in the menu in the top right corner, often represented by three dots (?) or lines ($\equiv$). \\
6  & - Klicken Sie auf diese Punkte oder Striche, um das Menü zu öffnen, und wählen Sie dann Einstellungen.                                            & - Click on these dots or dashes to open the menu and then select Settings.                                                                 & - Click on these dots or lines to open the menu, and then select Settings.                                                                     \\
   &                                                                                                                                                   &                                                                                                                                            &                                                                                                                                                \\
7  & Finden Sie die Option für die Startseite:                                                                                                         & Find the option for the \underline{home page}:                                                                                                         & Find the option for the \underline{homepage}:                                                                                                              \\
8  & - In den Einstellungen gibt es einen Abschnitt für die Startseite oder Beim Start.                                                                & - In Settings, there is a section for \underline{Home page} or On startup.                                                                             & \textbf{- In the settings, there is a section for the \underline{homepage} or start page.}                                                                 \\
9  & - Suchen Sie nach einer Option wie „Startseite festlegen“ oder „Beim Start“ (je nach Browser kann der genaue Wortlaut unterschiedlich sein).      & - Look for an option such as ‘Set \underline{home page}’ or ‘On startup’ (the exact wording may vary depending on your browser).                       & \textbf{- Look for an option like "Set \underline{homepage}" or "Start page" (the exact wording may vary depending on the browser).}                       \\
   &                                                                                                                                                   &                                                                                                                                            &                                                                                                                                                \\
10 & NZZ.de als Startseite festlegen:                                                                                                                  & Set NZZ.de as your \underline{start page}:                                                                                                             & \textbf{Set NZZ.de as your \underline{homepage}:}                                                                                                          \\
11 & - Wenn Sie die Option „Bestimmte Seite öffnen“ sehen, wählen Sie diese aus.                                                                       & - If you see the option ‘Open specific page’, select it.                                                                                   & - If you see an option called "Open a specific page," select it.                                                                               \\
12 & - Geben Sie in das Eingabefeld https://www.nzz.de ein.                                                                                            & - Enter https://www.nzz.de in the input field.                                                                                             & - Enter https://www.nzz.de in the input field.                                                                                                 \\
   &                                                                                                                                                   &                                                                                                                                            &                                                                                                                                                \\
13 & Einstellungen speichern:                                                                                                                          & Save settings:                                                                                                                             & Save settings:                                                                                                                                 \\
14 & - Die meisten Browser speichern die Änderung automatisch, sobald Sie das Feld verlassen.                                                          & - Most browsers save the change automatically as soon as you exit the field.                                                               & - Most browsers will save the change automatically as soon as you leave the field.                                                             \\
15 & Sie können das Einstellungsfenster nun schließen.                                                                                                 & You can now close the settings window.                                                                                                     & You can now close the settings window.                                                                                                        
\\ 
\bottomrule
\end{tabularx}}